\pdfoutput=1

\documentclass[11pt]{article}

\usepackage{emnlp2021}

\usepackage{times}
\usepackage{latexsym}
\usepackage{multirow}
\usepackage{graphicx}
\usepackage{algorithm}
\usepackage{algorithmic}
\usepackage{amsmath}
\usepackage{url}
\usepackage{comment}
\usepackage{enumitem}
\usepackage{color}
\usepackage{subfigure}
\usepackage[T1]{fontenc}
\usepackage{footnote}
\usepackage{hyperref}
\usepackage{listings}
\usepackage{booktabs}

\usepackage{xcolor}
\usepackage{threeparttable}

\definecolor{codegreen}{rgb}{0,0.6,0}
\definecolor{codegray}{rgb}{0.5,0.5,0.5}
\definecolor{codepurple}{rgb}{0.58,0,0.82}
\definecolor{backcolour}{rgb}{0.95,0.95,0.92}

\lstdefinestyle{mystyle}{
    backgroundcolor=\color{backcolour},   
    commentstyle=\color{codegreen},
    stringstyle=\color{codepurple},
    basicstyle=\ttfamily\scriptsize,
    breakatwhitespace=true,         
    breaklines=true,                 
    captionpos=b,                    
    keepspaces=true,                 
    numbers=none,                    
    numbersep=5pt,                  
    showspaces=false,                
    showstringspaces=false,
    showtabs=false,                  
    tabsize=2,
    columns=flexible,
    escapeinside={(*}{*)},
}

\usepackage{amssymb}
\usepackage[utf8]{inputenc}

\usepackage{microtype}

\title{YOLOX-PAI: An Improved YOLOX, Stronger and Faster than YOLOv6}

\author{Ziheng Wu$^1$\thanks{* Equal Contribution.}, Xinyi Zou$^{1*}$, Wenmeng Zhou$^{1}$, Jun Huang$^{1}$\\
$^1$ Platform of AI (PAI), Alibaba Group\\
\texttt{\{zhoulou.wzh, zouxinyi.zxy, wenmeng.zwm, huangjun.hj\}@alibaba-inc.com}
}

\begin{document}
\maketitle

\begin{figure*}[th]
	\centering
	\includegraphics[width=\linewidth]{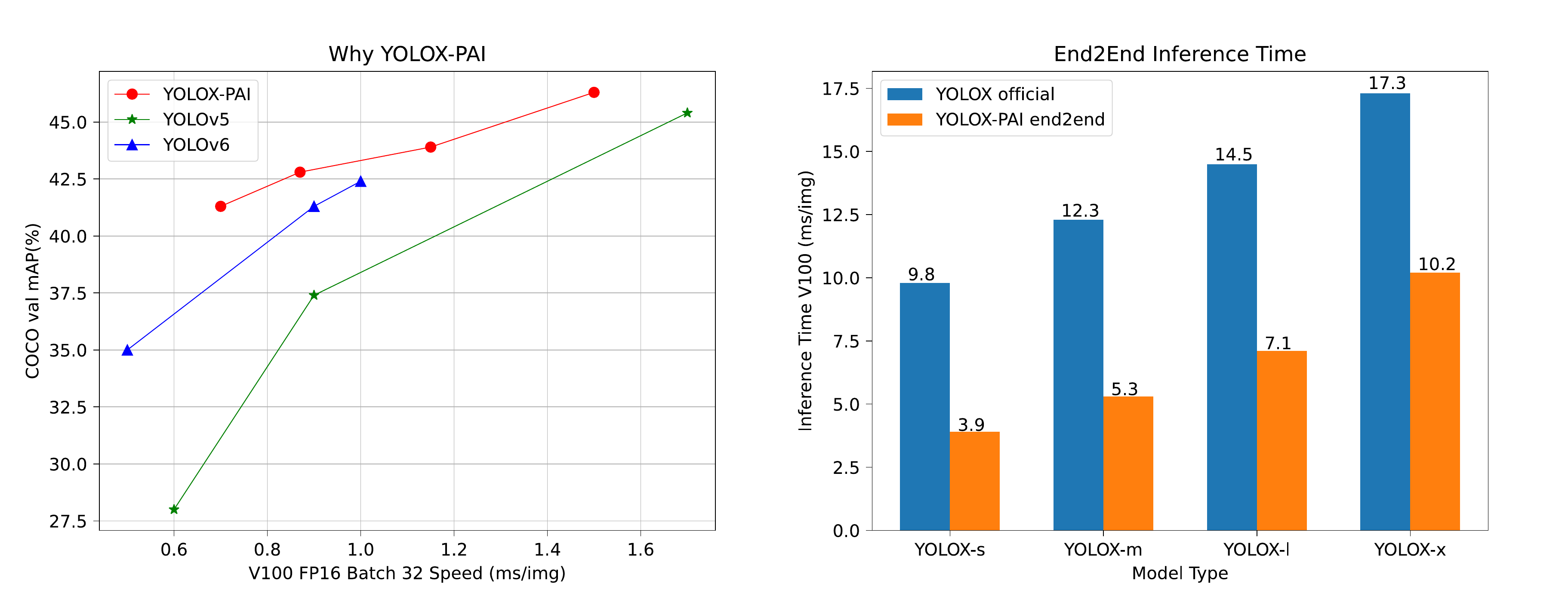}
	\caption{The comparisons between YOLOX-PAI and the existing methods.}\label{fig:result}
\end{figure*}

\begin{abstract}
	
We develop an all-in-one computer vision toolbox named EasyCV to facilitate the use of various SOTA computer vision methods. Recently, we add YOLOX-PAI, an improved version of YOLOX, into EasyCV. We conduct ablation studies to investigate the influence of some detection methods on YOLOX. We also provide an easy use for PAI-Blade which is used to accelerate the inference process based on BladeDISC and TensorRT. Finally, we receive 42.8 mAP on COCO dateset within 1.0 ms on a single NVIDIA V100 GPU, which is a bit faster than YOLOv6. A simple but efficient predictor api is also designed in EasyCV to conduct end2end object detection. Codes and models are now available at: \url{https://github.com/alibaba/EasyCV}.

\end{abstract}

\section{Introduction}

YOLOX \cite{ge2021} is one of the most famous one-stage object detection methods, and has been widely used in a various field, such as , defect inspection, \emph{etc}. It introduces the decoupled head and the anchor-free manner into the YOLO series, and receives state-of-the-art results among 40 mAP to 50 mAP. 

Considering its flexibility and efficiency, we intend to integrate YOLOX into our EasyCV, an all-in-one computer vision methods that helps even a beginner easily use a computer vision algorithm. In addition, we investigate the improvement upon YOLOX by using different enhancement of the detection backbone, neck, and head. Users can simply set different configs to obtain a suitable object detection model according to their own requirements. Also, based on PAI-Blade \footnote{https://github.com/alibaba/BladeDISC}(an inference optimization framework by PAI), we further speed up the inference process and provide an easy api to use PAI-Blade in our EasyCV. Finally, we design an efficient predictor api to use our YOLOX-PAI in an end2end manner, which accelerate the original YOLOX by a large margin. The comparisons between YOLOX-PAI and the state-of-the-art object detection methods have been shown in Fig.~\ref{fig:result}.

In brief, our main contributions are as follows:

\begin{itemize}
	\item We release YOLOX-PAI in EasyCV as a simple yet efficient object detection tool (containing the docker image, the process of model training, model evaluation and model deployment). We hope that even a beginner can use our YOLOX-PAI to accomplish his object detection tasks.
	
	\item We conduct ablation studies of existing object detection methods based on YOLOX, where only a config file is used to construct self-designed YOLOX model. With the improvement of the architecture and the efficiency of PAI-Blade, we obtain state-of-the-art object detection results among 40 mAP and 50 mAP within 1ms for model inference on a single NVIDIA Tesla V100 GPU.
	
	\item We provide a flexible predictor api in EasyCV that accelerates both the preprocess, inference and postprocess procedure, respectively. In this way, user can better use YOLOX-PAI for end2end objection detection task.
	
\end{itemize}

\section{Methods}

In this section, we will take a brief review of the used methods in YOLOX-PAI. We conduct several improvements on both the detection backbone, neck, head. We also use PAI-Blade to accelerate the inference process.

\subsection{Backbone}

Recently, YOLOv6 and PP-YOLOE \cite{xu2022pp} have replaced the backbone of CSPNet \cite{wang2020cspnet} to RepVGG \cite{ding2021repvgg}. In RepVGG, a 3x3 convolution block is used to replace a multi-branch structure during the inference process, which is beneficial to both save the inference time and improve the object detection results. Following YOLOv6,we also use a RepVGG-based backbone as a choice in YOLOX-PAI.

\subsection{Neck}

We use two methods to improve the performance of YOLOX in the neck of YOLOX-PAI, that is 1) Adaptively Spatial Feature Fusion (ASFF) \cite{liu2019learning} and its variance (denoted as ASFF\_Sim) for feature augmentation 2) GSConv \cite{li2022slim}, a lightweight convolution block to reduce the compute cost.

\begin{figure*}[th]
	\centering
	\includegraphics[width=\linewidth]{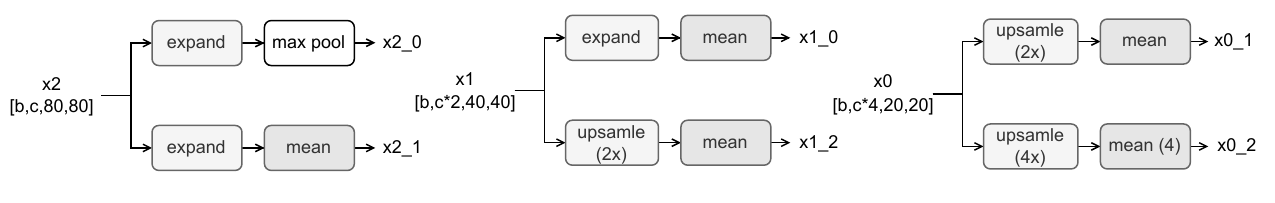}
	\caption{Operations in ASFF\_Sim. The expand operation is a slice operation based on the Focus layer.}
	\label{fig:asff_sim}
\end{figure*}

The original ASFF method uses several vanilla convolution blocks to first unify the dimension of different feature maps. Inspired by the Focus layer in YOLOv5, we replace the convolution blocks by using the non-parameter slice operation and mean operation to obtain the unified feature maps (denoted as ASFF\_Sim). To be specific, the operation for each feature map of the output of YOLOX is defined in Fig.~\ref{fig:asff_sim}.

\begin{figure}[th]
	\centering
	\includegraphics[width=\linewidth]{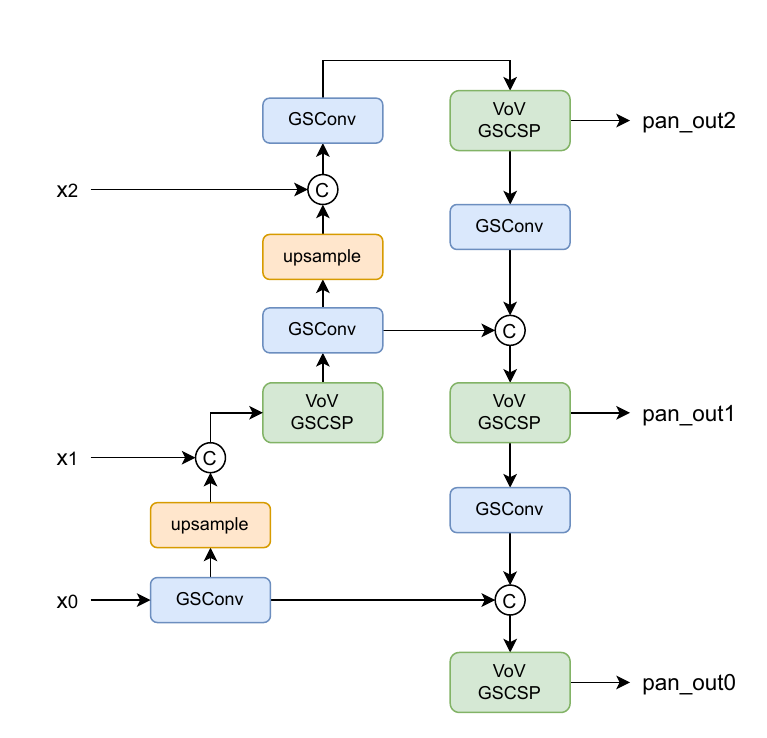}
	\caption{Architecture of the GSConv-based YOLOX neck.}\label{fig:gsconv}
\end{figure}

\begin{figure}[th]
	\centering
	\includegraphics[width=\linewidth]{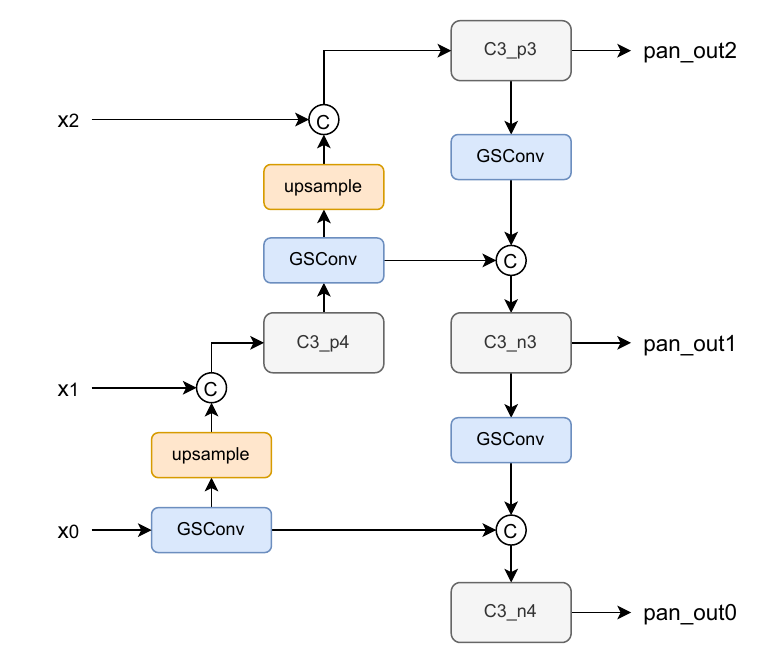}
	\caption{Architecture of the GSConv-based (part) YOLOX neck.}\label{fig:gsconv_part}
\end{figure}

We also use two kinds of GSConv-based neck to optimize YOLOX. The used neck architectures are shown in Fig~\ref{fig:gsconv} and Fig~\ref{fig:gsconv_part}. The differences of the two architectures are whether to replace all the blocks with GSConv. As proved by the authors, GSconv is specially designed for the neck where the channel reaches the maximum and the the size reaches the minimum.

\subsection{Head}

\begin{figure*}[th]
	\centering
	\includegraphics[width=\linewidth]{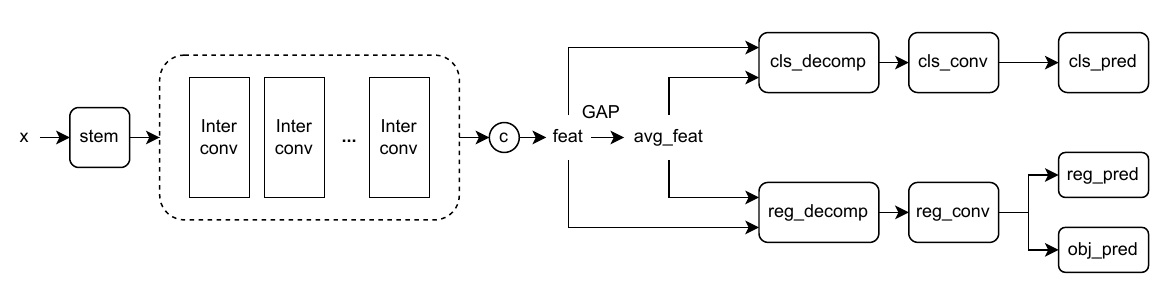}
	\caption{Architecture of the TOOD-Head.}\label{fig:head}
\end{figure*}

We enhance the YOLOX-Head with the attention mechanism as \cite{feng2021tood} to align the task of object detection and classification (denoted as TOOD-Head). The architecture is shown in Fig.~\ref{fig:head}. A stem layer is first used to reduce the channel, following by a group of inter convolution layers to obtain the inter feature maps. Finally, the adaptive weight is computed according to different tasks. We test the result of using the vanilla convolution or the repvgg-based convolution in the TOOD-Head, respectively.

\subsection{PAI-Blade}

PAI-Blade is an easy and robust inference optimization framework for model acceleration. It is based on many optimization techniques, such as Blade Graph Optimizer, TensorRT, PAI-TAO (Tensor Accelerator and Optimizer), and so on. PAI-Blade will automatically search for the best method to optimize the input model. Therefore, people without the professional knowledge of model deployment can also use PAI-Blade to optimize the inference process. We integrate the use of PAI-Blade in EasyCV so that users are allowed to obtain an efficient model by simply change the export config.

\subsection{EasyCV Predictor}

Along with the model inference, the preprocess function and the postprocess function are also important in an end2end object detection task, which are often ignored by the existing object detection toolbox. In EasyCV, we allow user to choose whether to export the model with the preprocess/postprocess procedure flexibly. Then, a predictor api is provided to conduct efficient end2end object detection task with only a few lines.

\section{Experiments}

\begin{table*}
	\centering
	\caption{Comparisons between YOLOX-PAI and SOTA methods on the COCO dataset.}
	\label{tab:sota}
	\begin{threeparttable}
		\begin{tabular}{l|ccccc} 
			\toprule[1.5pt]
			Model                 & Params (M) & Flops (G) & \begin{tabular}[c]{@{}l@{}}mAP\textsuperscript{val}\\\scriptsize{0.5:0.95 (672)}\end{tabular} & \begin{tabular}[c]{@{}l@{}}mAP\textsuperscript{val}\\\scriptsize{0.5:0.95 (640)}\end{tabular} & \begin{tabular}[c]{@{}l@{}}Speed\textsuperscript{V100}\\\scriptsize{fp16 bs32 (ms)}\end{tabular} \\ 
			\hline
			YOLOXs  $^\ddagger$               & 9.0        & 26.8      & 40.2                                                                            & 40.1     & 0.68                                             \\
			YOLOv6-tiny           & 15.0       & 36.7      & 41.3                                                                            & -        & 0.9                                              \\
			PAI-YOLOXs            & 15.9       & 36.8      & 41.5                                                                            & 41.4     & 0.70                                              \\
			YOLOv6-s              & 17.2       & 44.2      & 43.1                                                                            & 42.4     & 1.0                                              \\
			PAI-YOLOXs-ASFF       & 21.3       & 41.0      & 43.3                                                                            & 42.8     & 0.87                                             \\
			PAI-YOLOXs-ASFF-TOOD3 & 23.7       & 49.9      & 44.0                                                                            & 43.9     & 1.15                                             \\
			YOLOXm  $^\ddagger$               & 25.3       & 73.8      & 46.3                                                                            & 46.3     & 1.50                                             \\
			\bottomrule[1.5pt]
		\end{tabular}

		\begin{tablenotes}
			\footnotesize
			\item $^\ddagger$ denotes the re-implementation YOLOX result in EasyCV, which is optimized by PAI-Blade.
		\end{tablenotes}
	\end{threeparttable}
	
\end{table*}

\begin{table}
	\centering
	\caption{Ablation studies of ASFF and its variance.}
	
	\label{tab:ab_asff}
	\begin{threeparttable}
		\resizebox{\linewidth}{!}{
			\begin{tabular}{l|ccccc} 
				\toprule[1.5pt]
				
				Model                 & Params (M) & Flops (G) &  \begin{tabular}[c]{@{}l@{}}mAP\textsuperscript{val}\\\scriptsize{0.5:0.95 (640)}\end{tabular} & \begin{tabular}[c]{@{}l@{}}Speed\textsuperscript{V100}\\\scriptsize{fp16 bs32 (ms)}\end{tabular} \\ 
				\hline
				Baseline         & 15.9       & 36.8      & 41.4     & 0.70                                             \\
				+ASFF     & 21.3       & 41.0      & 42.8     & 0.87                                             \\
				+ASFF-Sim & 16.3       & 37.1      & 42.6     & 1.13                                             \\
				\bottomrule[1.5pt]
			\end{tabular}
		}
		\begin{tablenotes}
			\footnotesize
			\item * PAI-YOLOXs is the baseline.
		\end{tablenotes}
	\end{threeparttable}

\end{table}

\begin{table}
	\centering
	\caption{Ablation studies of GSConv block.}
	
	\label{tab:ab_gsconv}
	\begin{threeparttable}
		\resizebox{\linewidth}{!}{
			\begin{tabular}{l|ccccc} 
				\toprule[1.5pt]
				
				Model                 & Params (M) & Flops (G) &  \begin{tabular}[c]{@{}l@{}}mAP\textsuperscript{val}\\\scriptsize{0.5:0.95 (640)}\end{tabular} & \begin{tabular}[c]{@{}l@{}}Speed\textsuperscript{V100}\\\scriptsize{fp16 bs32 (ms)}\end{tabular} \\ 
				\hline
				Baseline         & 2.83       & 2.67      & 41.4     & 0.70                                             \\
				GSConv     & 1.22       & 1.33      & 41.6     & 0.78                                             \\
				GSConv\_part & 2.39       & 2.39      & 41.7     & 0.72                                             \\
				\bottomrule[1.5pt]
			\end{tabular}
		}
		\begin{tablenotes}
			\footnotesize
			\item * only the parameter and flops of neck is computed.
			\item * PAI-YOLOXs is the baseline.
		\end{tablenotes}
	\end{threeparttable}
	
\end{table}

\begin{table}
	\centering
	\caption{Ablation studies of TOOD-Head.}
	
	\label{tab:head}
	\begin{threeparttable}
		\resizebox{\linewidth}{!}{
			\begin{tabular}{l|ccccc} 
				\toprule[1.5pt]
				
				Model                 & Params (M) & Flops (G) &  \begin{tabular}[c]{@{}l@{}}mAP\textsuperscript{val}\\\scriptsize{0.5:0.95 (640)}\end{tabular} & \begin{tabular}[c]{@{}l@{}}Speed\textsuperscript{V100}\\\scriptsize{fp16 bs32 (ms)}\end{tabular} \\ 
				\hline
				Baseline         & 1.92       & 5.23      & 42.8     & 0.87                                             \\
				stack=2     & 2.22       & 7.71      & 43.5     & 1.09                                            \\
				stack=3 & 2.37       & 8.94      & 43.9     & 1.15                                             \\
				stack=4 & 2.53       & 10.18      & 44.1     & 1.24                                             \\
				stack=5 & 2.68       & 11.42      & 44.4     & 1.32                                             \\
				stack=6 & 2.83       & 12.66      & 44.7     & 1.40                                             \\
				\bottomrule[1.5pt]
			\end{tabular}
		}
		\begin{tablenotes}
			\footnotesize
			\item * only the parameter and flops of head is computed.
			\item * PAI-YOLOXs-ASFF is the baseline.
		\end{tablenotes}
	\end{threeparttable}
	
\end{table}

\begin{table}
	\centering
	\caption{End2end inference time of YOLOX-s with different export configs.}
	
	\label{tab:e2e}
	
	\resizebox{\linewidth}{!}{
		\begin{tabular}{c|c|c|c} 
			\toprule[1.5pt]
			
			export\_type                 & preprocess\_jit &use\_trt\_efficientnms    & \begin{tabular}[c]{@{}c@{}}Inference Time\textsuperscript{V100}\\\scriptsize{bs1 end2end (ms)}\end{tabular} \\ 
			\hline
			raw         & -       & -      & 24.58                                                 \\
			jit     & $\times$       & $\times$      & 18.30     \\
			jit		 & $\times$       & $\checkmark$      & 18.38       \\
			jit        & $\checkmark$      & $\times$    & 13.44    \\
			jit        &  $\checkmark$      &  $\checkmark$   & 13.04     \\
			blade     & $\times$       & $\times$      & 8.72     \\
			blade		 & $\times$       & $\checkmark$      & 9.39       \\
			blade        & $\checkmark$      & $\times$    & 3,93    \\
			blade        &  $\checkmark$      &  $\checkmark$   & 4.53     \\
			\bottomrule[1.5pt]
		\end{tabular}
	}

\end{table}

\begin{table}
	\centering
	\caption{Comparisons between end2end inference time of YOLOX-PAI and the official YOLOX (V100).}
	
	\label{tab:e2e}
	
	\resizebox{\linewidth}{!}{
		\begin{tabular}{c|c|c|c} 
			\toprule[1.5pt]
			
		Model                 & YOLOX-offical & YOLOX-PAI    & Speedup \\ 
			\hline
			YOLOX-s         & 9.8 ms      & 3.9 ms    & 251\%                                                 \\
			YOLOX-m     & 12.3 ms      & 5.3 ms    & 232\%    \\
			YOLOX-l		 & 14.5 ms      & 7.1 ms     & 204\%       \\
			YOLOX-x        & 17.3 ms     & 10.2 ms   & 170\%    \\
		
			\bottomrule[1.5pt]
		\end{tabular}
	}

\end{table}

In this section, we report the ablation study results of the above methods on the COCO dataset \cite{lin2014microsoft} and the comparisons between YOLOX-PAI and the SOTA object detection methods.

\subsection{Comparisons with the SOTA methods}

We select the useful improvements in YOLOX-PAI and compare it with the SOTA YOLOv6 method in Table~\ref{tab:sota}. It can be seen that YOLOX-PAI is much faster compared to the corresponding version of YOLOv6 with a better mAP (i.e., obtain +0.2 mAP and 22\% speed up, +0.2 mAP and 13\% speed up of the YOLOv6-tiny, and the YOLOv6-s model, respectively).

\subsection{Ablation studies}

\noindent  \textbf{Influence of Backbone.}
As shown in Table~\ref{tab:sota}, YOLOX with a RepVGG-based backbone achieve better mAP with only a little sacrifice of speed. It indeed adds more parameters and flops that may need more computation resource, but does not require much inference time. Considering its efficiency, we make it as a flexible config setting in EasyCV.

\noindent  \textbf{Influence of Neck.}
The influence of ASFF and ASFF\_Sim are shown in Table~\ref{tab:ab_asff}. It shows that, compared with ASFF, ASFF\_Sim is also benefical to improve the detection result with only a little gain of parameters and flops. However, the time cost is much larger and we will implement the CustomOP to optimize it in the future. The influence of GSConv is shown in Table~\ref{tab:ab_gsconv}. The result is that GSConv will bring 0.3 mAP and reduce 3\% of speed on a single NVIDIA V100 GPU.

\noindent  \textbf{Influence of Head.}
The influence of the TOOD-Head is shown in Table~\ref{tab:head}. We investigate the influence of different number of inter convolution layers. We show that when adding additional inter convolution layers, the detection results can be better. It is a trade-off between the speed and accuracy to choose a suitable hyperparameter. We also show that when replacing the vanilla convolution with the repconv-based convolution in the inter convolution layers, the result become worse. It will slightly improve the result when using repconv-based cls\_conv/reg\_conv layer (in Fig.~\ref{fig:head}) when the stack number is small (i.e., 2, 3).

\subsection{End2end results}

Table~\ref{tab:e2e} shows the end2end prediction results of YOLOXs model with different export configs. The keywords in the table are the same as in our EasyCV config file. It is evident that the blade optimization is useful to optimize the inference process. Also, the preprocess process can be greatly speed up by the exported jit model. As for the postprocess, we are still work on it to realize a better CustomOP that can be optimized by PAI-Blade for a better performance. In the right part of Fig~\ref{fig:result}, we show that with the optimization of PAI-Blade and EasyCV predictor, we can receive a satisfactory end2end inference time on YOLOX.

\section{Conclusion}

In this paper, we introduced YOLOX-PAI, an improved version of YOLOX based on EasyCV. It receives SOTA object detection results among 40 mAP and 50 mAP with the improvement of the model architecture and PAI-Blade. We also provide an easy and efficient predictor api to conduct end2end object detection in a flexible way. EasyCV is an all-in-one toolkit box that focuses on SOTA computer vision methods, especially in the field of self-supervised learning and vision transformer. We hope that users can conduct computer vision tasks immediately and enjoy CV by using EasyCV!

\section*{Acknowledgments}
We thank all the authors of the algorithms re-implemented in EasyCV for their contributions in the github community. We also thank Tianyou Guo, Nana Jiang, Haipeng Fang, Jiayu Chen and many other members of the Alibaba PAI team for their contribution and suggestions on building the YOLOX-PAI and EasyCV toolkit.

\bibliography{anthology}
\bibliographystyle{yolox-pai}

\end{document}